\documentclass[11pt]{article}
\usepackage{coling2016}
\usepackage{times}
\usepackage{url}
\usepackage{latexsym}
\usepackage{color,soul}
\usepackage{graphicx}
\usepackage{caption}
\usepackage{subcaption}
\usepackage{url}
\usepackage[T1]{fontenc}

\newcommand{\cc}{\mathbf{c}}
\newcommand{\xx}{\mathbf{x}}
\newcommand{\hh}{\mathbf{h}}

\title{A Character-level Convolutional Neural Network\\for Distinguishing Similar Languages and Dialects}

\author{Yonatan Belinkov \and James Glass \\
 MIT
  Computer Science and Artificial Intelligence Laboratory \\
  Cambridge, MA 02139, USA \\
  {\tt \{belinkov, glass\}@mit.edu}  }

\date{}

\begin{document}
\maketitle
\begin{abstract}
Discriminating between closely-related language varieties is considered a challenging and important task. This paper describes our submission to the DSL 2016 shared-task, which included two sub-tasks: one on discriminating similar languages and one on identifying Arabic dialects.  We developed a character-level neural network for this task. Given a sequence of characters, our model embeds each character in vector space, runs the sequence through multiple convolutions with different filter widths, and pools the convolutional representations to obtain a hidden vector representation of the text that is used for predicting the language or dialect. We primarily focused on the Arabic dialect identification task and obtained an F1 score of 0.4834, ranking 6th out of 18 participants. 
We also analyze errors made by our system on the Arabic data in some detail, and point to challenges such an approach is faced with.\footnote{The code for this work is available at \url{https://github.com/boknilev/dsl-char-cnn}.}
\end{abstract}

\section{Introduction}
\label{intro}

Automatic language identification is an important first step in many applications. For many languages and texts, this is a fairly easy step that can be solved with familiar methods like n-gram language models. However, distinguishing between similar languages is not so easy. The shared-task on discriminating between similar languages (DSL) has offered a test bed for evaluating models on this task since 2014~\cite{zampieri:2014:VarDial,zampieri:2015:LT4VarDial}. The 2016 shared-task included two sub-tasks: (1) discriminating between similar languages from several groups; and (2) discriminating between Arabic dialects~\cite{dsl2016}. The following language varieties were considered in sub-task 1:
 Bosnian, Croatian, and Serbian; 
Malay and Indonesian;
Portuguese of Brazil and Portugal;
Spanish of Argentina, Mexico, and Spain;
and French of France and Canada. In sub-task 2, the following Arabic varieties were considered:
Levantine, Gulf, Egyptian, North African, and  Modern Standard Arabic (MSA). 

The training datasets released in the two sub-tasks were very different. Sub-task 1 was comprised of journalistic texts and included  training and development sets. Sub-task 2 had automatic transcriptions of spoken recordings and included only a training set. This was the first time DSL has offered a task on Arabic dialects. The shared-task also included an open track that allows additional resources, but we have only participated in the closed track.

Previous DSL competitions attracted a variety of methods, achieving very high results with accuracies of over 95\%. Most teams used character and word n-grams with various classifiers. One team in 2015 used vector embeddings of words and sentences~\cite{francosalvador-rosso-rangel:2015:LT4VarDial}, achieving very good, though not state-of-the-art results. They trained unsupervised vectors and fed them as input to a classifier. Here we are interested in \textit{character-level} neural network models. Such models showed recent success in other tasks~\cite[among many others]{KimAAAI1612489}.  The basic question we ask is this: how well can a character-level neural network perform on this task without the notion of a word? To answer this, our model takes as input a sequence of characters, embeds them in vector space, and generates a high-level representation of the sequence through multiple convolutional layers. At the top of the network, we output a probability distribution over labels and backpropagate errors, such that the entire network can be learned end-to-end. 

We experimented with several configurations of convolutional layers, focusing on Arabic dialect identification (sub-task 2). We also participated in sub-task 1, but have not tuned our system to this scenario. Our best system obtained an F1 score of 0.4834 on the Arabic sub-task, ranking 6th out of 18 participants. The same system did not perform well on sub-task 1 (ranked 2nd to last), although we have not spent much time on adapting it to this task.
In the next section, we discuss related work on identifying similar languages and dialects. We then present our methodology and results, and conclude with a discussion and a short error analysis for the Arabic system that sheds light on potential sources of errors.

\section{Related Work}

Discriminating similar languages and dialects has been the topic of two previous iterations of the DSL task~\cite{zampieri:2014:VarDial,zampieri:2015:LT4VarDial}. The previous competitions proved that despite very good performance (over 95\% accuracy), it is still a non-trivial task that is considerably more difficult than identifying unrelated languages. Admittedly, even humans have a hard time identifying the correct language in certain cases, as observed by \newcite{GOUTTE16.69}. The previous shared-task reports contain detailed information on the task, related work, and participating systems. Here we only highlight a few relevant trends.

In terms of features, the majority of the groups in previous years used sequences of characters and words. A notable exception is the use of word white-lists by \newcite{porta-sancho:2014:VarDial}. Different learning algorithms were used for this task, most commonly linear SVMs or maximum entropy models. Some groups formulated a two-step classification model: first predicting the language group and then predicting an individual language. For simplicity, we only trained a single multi-class classification model, although we speculate that a two-step process could improve our results. For  Arabic dialect identification  (sub-task 2), one could first distinguish MSA from all the other dialects, and then identify the specific dialect. 

Last year, \newcite{francosalvador-rosso-rangel:2015:LT4VarDial} used vector embeddings of words and sentences, achieving very good results, though not state-of-the-art. They trained unsupervised word vectors and fed them as input to a classifier. In contrast, we build an end-to-end neural network over \textit{character} sequences, and train character embeddings along with other parameters of the network. Using character embeddings is particularly appealing for this task given the importance of character n-gram features in previous work. In light of the recent success of character-level neural networks in various language processing and understanding tasks~\cite{santos2014learning,NIPS2015-5782-Zhang,luong2016achieving,KimAAAI1612489},
we were interested to see how far one can go on this task without any word-level information.

Finally, this year's task offered a sub-task on Arabic dialect identification. It is unique in that the texts are automatic transcriptions generated by a speech recognizer. Previous work on Arabic dialect identification mostly used written texts~\cite{Zaidan:2014:ADI:2645242.2645248,malmasi-et-al:2015:adi} or speech recordings, with access to the acoustic signal~\cite{biadsy-hirschberg-habash:2009:Semitic,Ali+2016}. For example, \newcite{Ali+2016} exploited both acoustic and ASR-based features, finding that their combination works best. Working with automatic transcriptions obscures many dialectal differences (e.g. in phonology) and leads to inevitable errors. Still, we were interested to see how well a character-level neural network can perform on this task, without access to acoustic features.  

\section{Methodology}

We formulate the task as a multi-class classification problem, where each language (or dialect) is a separate class. We do not consider two-step classification, although this was found useful in previous work~\cite{zampieri:2015:LT4VarDial}. 
Formally, given a collection of texts and associated labels, $\{t^{(i)}, l^{(i)}\}$, we need to find a predictor $f: t \rightarrow l$. Our predictor is a neural network over character sequences. 
Let $ t := \cc = c_1, \cdots, c_L $ denote a sequence of characters, where $L$ is a maximum length that we set empirically. Longer texts are truncated and shorter ones are padded with a special \texttt{PAD} symbol. Each character $c$ in the alphabet is represented as a real-valued vector $x_c$. This character embedding is learned during training. 

\pagebreak
Our neural network has the following structure:
\begin{itemize}
\item Input layer: mapping the character sequence  $\cc$ to a vector sequence $\xx$. The embedding layer is followed by dropout. 
\item Convolutional layers: multiple parallel convolutional layers, mapping the vector sequence $\xx$ to a hidden sequence $\hh$. Each convolution is followed by a Rectified Linear Unit (ReLU) non-linearity~\cite{AISTATS2011_GlorotBB11}. The outputs of all the convolutional layers are concatenated. 
\item Pooling layer: a max-over-time pooling layer, mapping the vector sequence $\hh$ to a single hidden vector $h$ representing the entire sequence. 
\item Fully-connected layer: one hidden layer with a ReLU non-linearity and dropout, mapping $h$ to the final vector representation of the text, $h'$.
\item Output layer: a softmax layer, mapping $h'$ to a probability distribution over labels $l$. 
\end{itemize}

During training, each sequence is fed into this network to create label predictions. As errors are back-propagated down the network, the weights at each layer are updated, including the embedding layer. During testing, the learned weights are used in a forward step to compute a prediction over the labels. We always take the best predicted label for evaluation.

\subsection{Training details and submitted runs}
We train the entire network jointly, including the embedding layer. We use Adam~\cite{kingma2014adam} with the default original parameters to minimize the cross-entropy loss. Training is run with shuffled mini-batches of size 16 and stopped once the loss on the dev set stops improving; we allow a patience of 10 epochs. Our implementation is based on Keras~\cite{chollet2015keras} with the TensorFlow backend~\cite{tensorflow2015-whitepaper}.

We mostly experimented with the sub-task 2 dataset of Arabic dialects. Since the official shared-task did not include a dedicated dev set, we randomly allocated 90\% of the training set for development. We tuned the following hyperparameters using this split (chosen parameters are in bold): embedding layer dropout (\textbf{0.2}, 0.5), fully-connected layer dropout (0.2, \textbf{0.5}), maximum text length (200, \textbf{400}, 800), character embedding size (25, \textbf{50}, 100), fully-connected layer output size (100, \textbf{250}). Removing the fully-connected layer led to a small drop in performance.

For the convolutional layers, we experimented with different combinations of filter widths and number of filters. We started with a single filter width and noticed that a  width of 5 characters performs fairly well with enough filters (250). We then added multiple widths, similarly to a recent character-CNN used in language modeling~\cite{KimAAAI1612489}. Using multiple widths led to a small improvement. Our best configuration was: $\{1*50, 2*50, 3*100, 4*100, 5*100, 6*100, 7*100 \}$, where $w*n$ indicates $n$ filters of width $w$. 

Since the shared-task did not provide a dev set for sub-task 2, we explored several settings in our three submitted runs. \textbf{Run 1} used 90\% of the training set for training and 10\% for development. \textbf{Run 2} used 100\% of the training for training, but stopped at the same epoch as Run 1. \textbf{Run 3} used 10 different models, each trained on a different random 90\%/10\% train/dev split, with a plurality vote among the 10 models to determine the final prediction.  

For the (larger) dataset of sub-task 1, we did not perform any tuning of  hyperparameters and used the same setting as in sub-task 2, except for a larger mini-batch size (64) to speed up training. This was our \textbf{Run 1}, whereas in \textbf{Run 2} we used more filter maps, following the setting in \cite{KimAAAI1612489}: $\{1*50, 2*100, 3*150, 4*200, 5*200, 6*200, 7*200 \}$. \textbf{Run 3} was the same as Run 2, but with more hidden units (500) and a higher dropout rate (0.7) in the fully-connected layer.   

\section{Results}
\label{sec:results}
Table~\ref{tab:results-all} shows the results of our submitted runs, along with two baselines: a random baseline for sub-task 1, test set A (a balanced test set), and a majority baseline for sub-task 2, test set C (a slightly unbalanced test set). We also report results of the best performing systems in the shared-task.  

In sub-task 2, test set C, we notice a fairly large difference between our runs. Our best result, with Run 3, used plurality voting among 10 different models trained on 90\% of the training data. We chose this strategy in an effort to avoid overfitting. However, during development we obtained accuracy results of around 57-60\% on a separate train/dev split, so we suspect there was still  overfitting of the training set. With this run our team ranked 6th out of 18 teams according to the official evaluation. 

For sub-task 1, test set A, larger models perform somewhat better, due to the larger training set. In this case we only have a small drop of about 2\% in comparison to our performance on the dev set (not shown). Our system did not perform very well compared to other teams (we ranked 2nd to last), but this may be expected as we did not tune our system for this dataset.

\begin{table}[t]
\center
\begin{tabular}{|lllllll|}
\hline
\bf Test Set & \bf Track & \bf Run & \bf Accuracy & \bf F1 (micro) & \bf F1 (macro) & \bf F1 (weighted) \\ 
\hline
A & closed & Baseline & 0.083 & & & \\
\hline
A & closed & run1 & 0.8042 & 0.8042 & 0.8017 & 0.8017 \\
A & closed & run2 & 0.825 & 0.825 & 0.8249 & 0.8249 \\
A & closed & run3 & \bf 0.8307 & \bf 0.8307 & \bf 0.8299 & \bf 0.8299 \\
\hline
A & closed & Best & \it 0.8938 & & & \it 0.8938 \\ 
\hline
\hline
C & closed & Baseline & 0.2279 & & & \\
\hline
C & closed & run1 & 0.4487 & 0.4487 & 0.4442 & 0.4449 \\
C & closed & run2 & 0.4357 & 0.4357 & 0.4178 & 0.4181 \\
C & closed & run3 & \bf 0.4851 & \bf 0.4851 & \bf 0.4807 & \bf 0.4834 \\
\hline
C & closed & Best & \it 0.5117 & & & \it 0.5132 \\
\hline
\end{tabular}
\caption{Results for our submitted runs. Best results out of our runs are in \textbf{bold}; best overall systems shown in \textit{italics} for reference. Refer to the text for a description of runs and baselines.}
\label{tab:results-all}
\end{table}

\begin{figure}
\centering
\begin{subfigure}[t]{0.45\textwidth}
  \includegraphics[width=\textwidth]{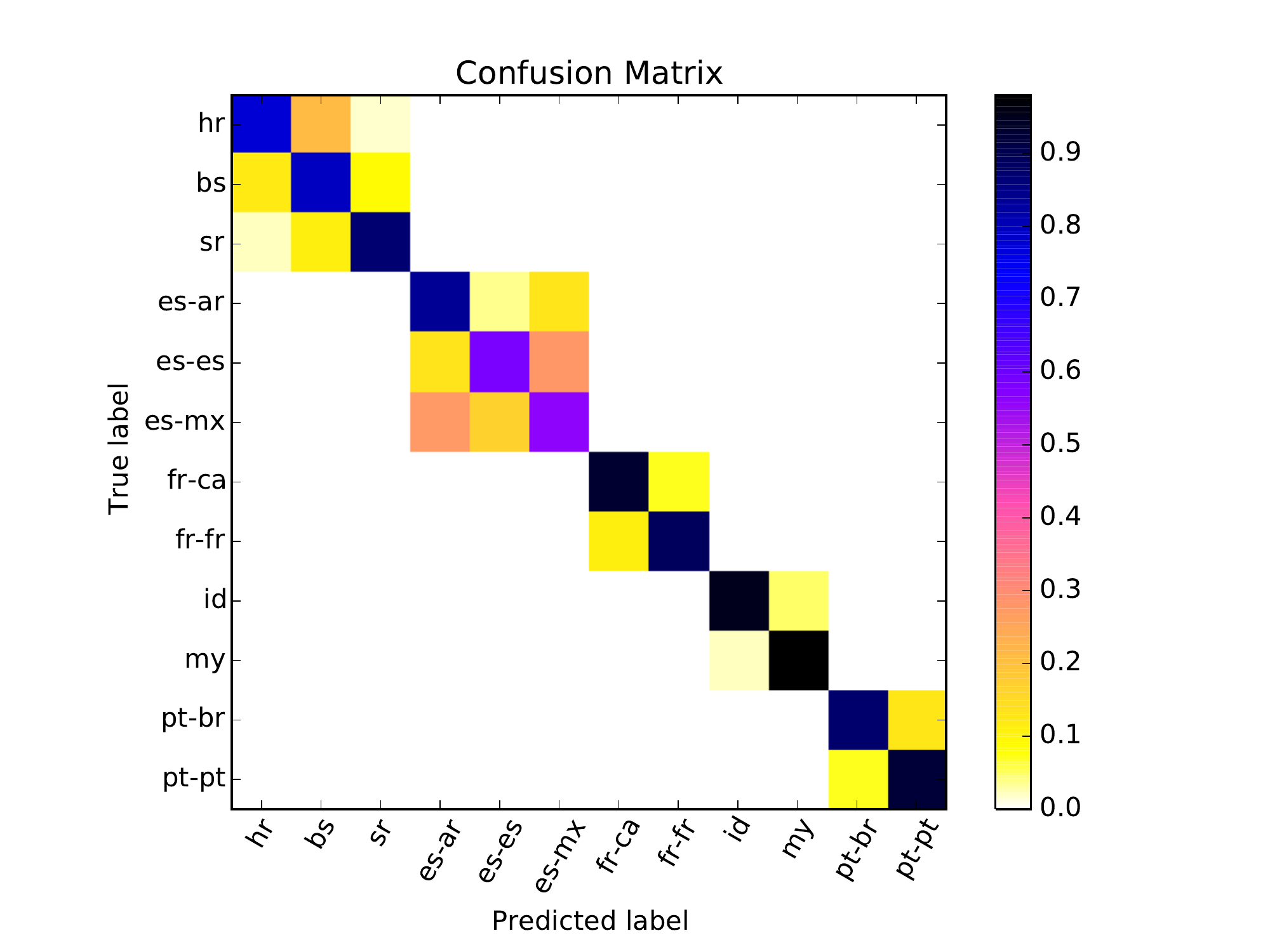}
  \caption{Task 1, test set A, Run 3.}
  \label{fig:a-3}
\end{subfigure}
\begin{subfigure}[t]{0.45\textwidth}
  \includegraphics[width=\textwidth]{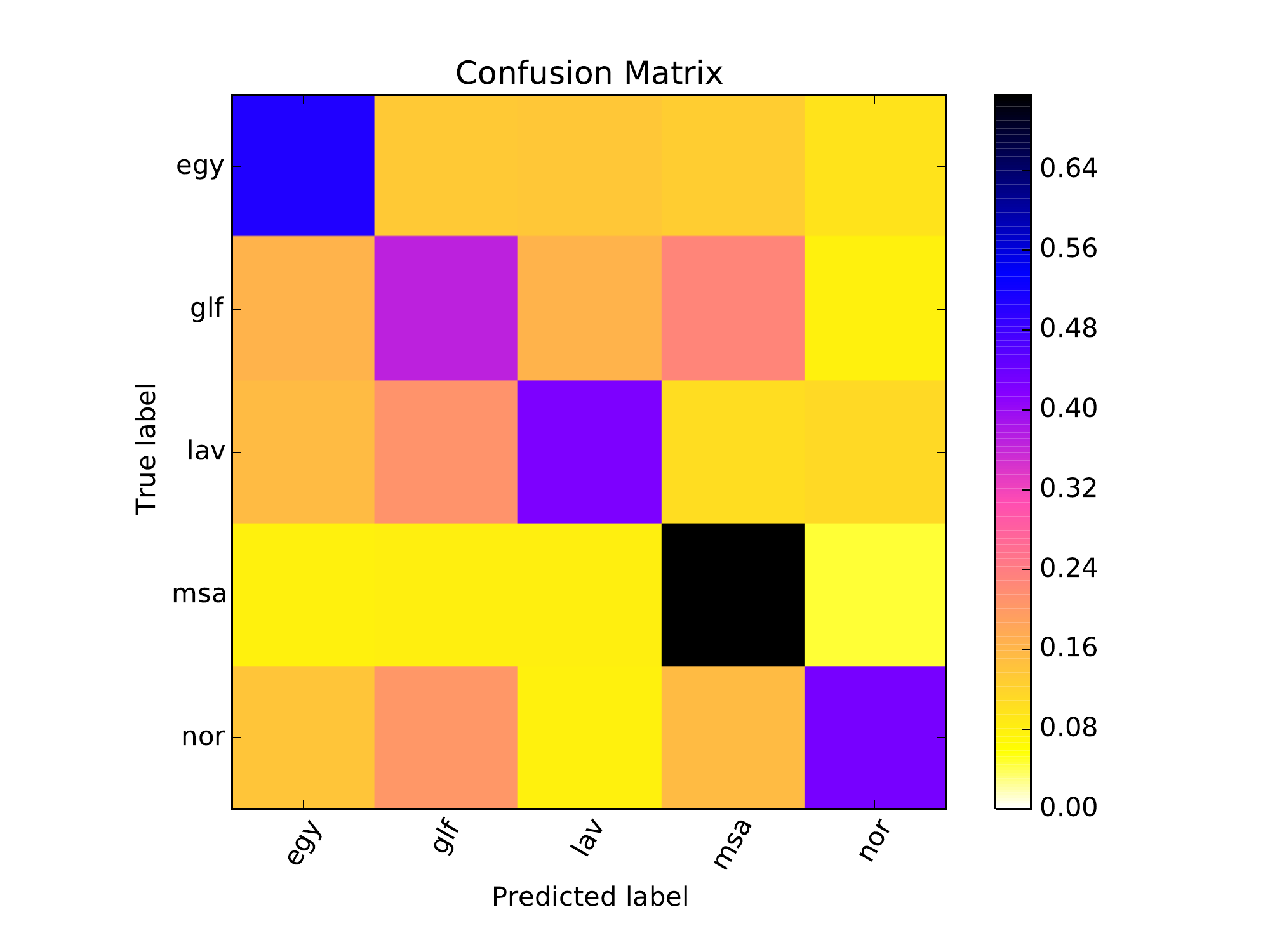}
  \caption{Task 2, test set C, Run 3.}
  \label{fig:c-3}
\end{subfigure}
\caption{Confusion matrices for our best runs on test sets A and C. Best viewed in color.}
\label{fig:confusion}
\end{figure}

Figure~\ref{fig:confusion} shows confusion matrices for our best runs. Clearly, the vast majority of the confusion in sub-task 1 (test set A) comes from languages in the same group, whereas languages from different groups are rarely confused. The Spanish varieties are the most difficult to distinguish, with F1 scores between 0.57 (Mexican Spanish) and 0.75 (Argentinian Spanish). The South Slavic languages are less confused, with F1 scores of 0.75-0.83. French and Portuguese languages are easier to distinguish (F1 around 0.90), and Malay and Indonesian are the least confused (F1 of 0.96-0.97). 

Turning to sub-task 2 (test set C), we see much more confusion, as also reflected in the final results (Table~\ref{tab:results-all}). Gulf is the most confusing dialect: true Gulf examples are often predicted as MSA, and true Levantine and North African examples are wrongly predicted as Gulf relatively frequently. This is also reflected in a low F1 score of 0.34 for Gulf. The other dialects have higher F1 scores ranging between 0.47 and 0.50, with MSA having an F1 of 0.60, making it the easiest variety to detect.

\section{Discussion}
In this section we focus on the Arabic dataset (sub-task 2, test set C) and consider some of our findings. The first issue that should be stressed is the nature of the data. As the texts are automatically generated speech transcriptions, they contain mistakes that depend on the training data used for the speech recognizer. This might have a negative effect on the ability to correctly recognize the dialect just from the transcriptions. For comparison,  previous work found that using acoustic features improves dialect recognition in a similar setting~\cite{Ali+2016}. Secondly, the transcribed texts use a transliteration system\footnote{\url{http://www.qamus.org/transliteration.htm}.} that was designed for written MSA and obscures many dialectal features that are important for distinguishing  Arabic dialects, especially phonological features.

The fact that MSA is confused with dialectal varieties at all may sound surprising at first. However, due to the writing scheme many of the known differences between MSA and the dialects are not accessible. In addition, MSA is often mixed with dialectal Arabic in many settings (e.g. news broadcasts), so it is reasonable to find MSA features in dialectal speech. 

Levantine and Gulf dialects are relatively close geographically and linguistically, so confusing one with the other might be expected. The confusion between Gulf and North African dialects is more surprising, although there are always parallel innovations in dialects that stem from a mutual ancestor, as all Arabic dialects do. 

Other factors that can play into similarities and differences are local foreign languages, religious terminology, and genre and domain. However, these are aspects that are difficult to tease apart without a more careful analysis, and they also depend on the data sources and the ASR system. 

\paragraph{Error analysis}
We conclude this discussion with an analysis of several example errors made by our systems, as shown in Figure~\ref{fig:errors}. These are examples from a dev set that we kept held out during the development of our system. In each case, we give the true and predicted labels, the text, and a rough translation. Note that the translations are often questionable due to ASR mistakes and the lack of context. 

In the first example, an MSA text is predicted as Levantine, possibly due to the word \textit{AllbnAnyp} ``Lebanese", whose character sequence could be typical of Levantine texts. There are clear MSA features like the dual forms \textit{hmA} and \textit{-An}, but these are ambigious with dialectal forms that are found in the training data. Note also the Verb-Subject word order, typical of MSA -- such a pattern requires syntactic knowledge and cannot be easily detected with simple character (or even word) features. 

The second error shows an example of a mixed text, containing both dialectal (\textit{>h HDrp} ``yes the honorable", \textit{bdy} ``I want") and MSA features (\textit{hl syHdv ``will it happen"}). It is an example of mixing that is common in spoken data and can confuse the model. 

In the third example, the form \textit{Alm\$kwk fyh} ``doubtful" has a morphological construction that is typical of MSA. Such a feature is difficult for a character-model to capture and might require morphological knowledge.

In the fourth example, the words \textit{AlmAlky} and \textit{AlhA\$my} are actually seen most commonly in MSA in the training data, but they do not contain typical character-sequences. The phrase \textit{Al Al}, which indicates some stuttering, is more common in some of the dialects than in MSA in the training data, but is about as common in NOR and MSA. 

In the fifth example, the word \textit{byt>vr} ``influences" is likely misrecognized, as in Egyptian it would probably be \textit{byt>tr}, but an Arabic language model with much MSA in it might push the ASR system towards the sequence \textit{>vr}, which in turn is more common in Gulf. 

In the seventh example, the phrase \textit{<HnA wyAhm} ``we're with them" might be more indicative of Gulf, but in any case is rare in the training data in both North African and Gulf. \textit{bqyt} ``remained'' should have been a good clue for North African, and indeed it appears 5 times in training as North African and not at all as Gulf. 

\pagebreak

\section{Conclusion}
In this work, we explored character-level convolutional neural networks for discriminating between similar languages and dialects. We demonstrated that such a model can perform quite well on Arabic dialect identification, although it does not achieve state-of-the-art results. We also conducted a short analysis of errors made by our system on the Arabic dataset, pointing to challenges that such a model is faced with. 

A natural extension of this work is to combine word-level features in the neural network. White-lists of words typical of certain dialects might also help, although in preliminary experiments we were not able to obtain performance gains by adding such features. We are also curious to see how our model would perform with access to the speech recordings, for example by running it on recognized phone sequences or directly incorporating acoustic features. This, however, would require a different preparation of the dataset, which we hope would be made available in future DSL tasks.

\begin{figure}
\footnotesize
\centering
\begin{tabular}{l p{1cm} p{1cm} p{6cm} p{6cm}}
& True & Predicted & Text & Translation \\
\hline
1 & MSA & Levantine & AndlEt AlHrb AllbnAnyp EAm 1975 >Syb bxybp >ml whmA yrwyAn kyf ynhArwn wqthA & "The Lebanon war erupted in 1975, he was disappointed and they both tell how they deteriorated back then" \\
2 & MSA & Egyptian & >h HDrp AlEmyd AlAHtkAk bdy dm\$q AlEASmp AlsyAdyp AlEASmp AlsyAsyp fy fy >kvr mn mrp wbEmlyp nwEyp kbyrp jdA hl syHdv AlmnErj fy h*h AlmwAjhp & "Yes, the honorable general, the friction, I want Damascus the sovereign capitol the sovereign capitol more than once and in a very large high-quality operation, will the turn take place in this confrontation" \\
3 & MSA & Gulf & >mA xrjt ElY tlfzywn Aldwlp fy Alywm AltAly lvwrp wqlt lh HAfZ ElY tAryx >ql AlwzArp Alywm Thr AlbrlmAn mn AlEDwyAt Alm\$kwk fyh <dY msyrp Al<SlAH wbdA h*A qbl SlAp AljmEp & But (I) went on the state television the following day after the revolution and told him, keep the history at least the ministry today, cleanse the parliament from the doubtful memberships if(?) the course of reform and this started before the Friday prayer"  \\
4 & MSA & North African & >wlA Al Al Alsyd AlmAlky ytmnY mn TArq AlhA\$my Alxrwj wlA yEwd & "First, mister Al-Maliki wants Tariq Al-Hashimi to exit and not return" \\
5 & Egyptian & Gulf & >nA bEmrnA ftrp mn HyAty snp Al>xyrp mtEwd ElY wqf Altfrd AHtlAly tEtbr llmsjd gryb mn mjls AlwzrA' wmjls Al\$Eb wAl\$wrY wnqAbp AlmHAmyn wxlAfh fkAn >y wAlAHtjAjyp byt>vr bhA Almsjd b\$>n Al>wDAE & "I in my life, time in my life, last year, used to stop being alone of occupation(?) that is considered for a mosque close to the cabinet and the parliament and the council and the bar association and behind it, and the protest, the mosque influences it because of the situation" \\
6 & Gulf & North African & lxwAtm Sgyrp fy AlHjm lkn tHtAj lEnAyp xASA \$mAtp ksArp wHtY AlxSm ArtHnA ElyhA bn\$wf h*A Altqryr wnErf mn xlAlh >kvr En tktlAt Alywm bfqr Aldm AlsyAsAt mE Alzmyl lwrA & "The rings/stamps are of small size but they need special care malice(?) breaker(?) and even the discount, we are happy with it, we see this report and through it we know more about the blocks/fights of today in lack of blood, the policies with the colleague are backwards" \\
7 & North African & Gulf & "\$Ahd tglb wAjb |xr mr Ebr EddA mn brnAmjh <HnA wyAhm lA ymnE xrwj bAlb\$r ftzydh whlA Em lxrwqAt AlHq Al\$yx xAld Hqq mEy lA yglq fyjb hdf AlnAtw bAlAxtyAr mn Altwqf lA tqAs bqyt Endk nsmH lkl \$y' HtY tqrr trHb & "See a must win last time through some of his programs, we are with them, will not prevent leaving with someone, and add more to it, and are now the violations, the truth the Sheikh Khalid interrogated me, he is not worried, and the goal of NATO must be to choose to stop, cannot be compared, still with you, I will permit everything until deciding to welcome"   \\
\end{tabular}
\caption{Example errors made by our system on the Arabic data set (sub-task 2). Some translations are questionable due to quality of transcription and lack of context.}
\label{fig:errors}
\end{figure}

\section*{Acknowledgments}
The authors would like to thank Sergiu Nisioi and Noam Ordan for helpful discussions. 
This work was supported by the Qatar Computing Research Institute (QCRI). Any opinions, findings, conclusions, or recommendations expressed in this paper are those of the authors, and do not necessarily reflect the views of the funding organizations.

\bibliographystyle{acl}
\bibliography{DSL} % you bib file(s) go here 

\end{document}